# Investigating the stylistic relevance of adjective and verb simile markers

**Suzanne Mpouli**
UPMC, LIP6, Paris
mpouli@acasa.lip6.fr

**Jean-Gabriel Ganascia**
UPMC, LIP6, Paris
jean-gabriel.ganascia@lip6.fr

## 1 Introduction

Similes are figures of speech in which the similarities as well as the differences between two or more semantically unrelated entities are expressed by means of a linguistic unit. This unit, also called marker, can either be a morpheme, a word or a phrase. Since similes rely on comparison, they occur in several languages of the world. Depending on the marker used and of the semantic or structural nuances it introduces, two main simile classifications have been proposed. From her study of images built on figures of resemblance in Baudelaire's poetry, Bouverot (1969) distinguishes between type I and type II similes. Whereas type I similes are denoted by a finite number of comparatives, prepositions or conjunctions which traditionally convey a comparison, type II similes are observed after a verb or an adjective which semantically contains the idea of similitude or difference. Leech and Short (2007) propose a stricter distinction and separate conventional similes of the form "X is like Y" from quasi-similes which revolve around all other linguistic constructions. This partition seems to take into account the fact that some simile markers are often preferred in a particular language. For example, 'like' in English or 'comme' in French are often presented as the prototypical simile markers and are more frequently used than the other markers.

Similes play an important role in literary texts not only as rhetorical devices and as figures of speech but also because of their evocative power, their aptness for description and the relative ease with which they can be combined with other figures of speech (Israel et al. 2004). Detecting all types of simile constructions in a particular text therefore seems crucial when analysing the style of an author. Few research studies however have been dedicated to the study of less prominent simile markers in fictional prose and their relevance for stylistic studies. The present paper studies the frequency of adjective and verb simile markers in a corpus of British and French novels in order to determine which ones are really informative and worth including in a stylistic analysis. Furthermore, are those adjectives and verb simile markers used differently in both languages?

## 2 Adjective and verb simile markers

Comparison being a semantic construction, there exist no comprehensive and finite lists of verb and adjective simile markers. The choice of the adjective and verb simile markers used in this experiment has been based on their ability to introduce phrasal similes, i.e. similes in which the compared term is a common noun. As far as verb markers are concerned, were ignored impersonal constructions which only accept indefinite pronouns as subjects such as in the French sentence "J'aime sa voix, on eût dit une pluie de grelots".

Under the category of verb simile markers, were included modified forms not found in the literature such as 'be/become….kind/sort/type of' and 'verb+ less/more than' as well as their respective French equivalents 'être/devenir….espèce/type/genre/sorte de' and 'verb+ plus/moins que'. As a matter of fact, even though expressions such as 'kind of' or 'sort of' are often cited among metaphorical markers (Goatly 1997), they were judged not specific enough to explicitly introduce a comparison on their own. In the case of the comparison markers 'less than' and 'more than', the issue rather lies in the fact that they are also used in other types of constructions and conjoining them to a verb form seems a good compromise to restrict their polysemy.

In English, in addition to adjectives conveying the idea of comparison, adjective simile markers include adjectives formed by adding the suffix '-like' and by joining a noun to an adjective of colour. Those two types of adjectives are particularly interesting stylistically speaking because they can potentially be used to create neologisms. Table 1 lists all the adjective and simile markers used for the experiment.

|  | **Verbs** | **Adjectives** |
|---|---|---|
| **English** | *resemble, remind, compare, seem, verb + less than, verb + more than, be/become… kind/sort/type of* | *similar to, akin to, identical to, analogous to, comparable to, compared to, reminiscent of, -like, noun+colour* |
| **French** | *ressembler à, sembler, , rappeler, faire l'effet de, faire penser à, faire songer à, donner l'impression de, avoir l'air de, verb + plus que, verb + moins que, être/devenir…espèce/type/ genre/sorte de* | *identique à, tel, semblable à, pareil à, similaire à, analogue à, égal à, comparable à* |

Table 1. Adjective and verb simile markers

## 3 Corpus building and extraction method

To extract simile candidates, digital versions of various literary texts were collected mainly from the Project Gutenberg website[1] and from the Bibliothèque électronique du Québec[2], for British and French novels respectively. Most of the novels included in the corpus were written during the 19th century so as to ensure linguistic homogeneity and because that century witnessed the novel imposing itself as a predominant literary genre. By observing a ration of least 3 novels per writer, we were able to put together a corpus of 1191 British novels authored by 62 writers and a corpus of 746 French penned by 57 novelists

Each corpus was tagged and chunked using TreeTagger, a multilingual decision tree part-of-speech tagger and a syntactic chunker (Schmid, 1994). The output of the tagger was further used to determine sentence boundaries. Each extracted sentence is considered a simile candidate if it contains one of the marker immediately followed by a noun-headed noun phrase.

## 4 Results

Since similes are realised first and foremost within a sentence, simile frequency was first calculated by dividing the number of occurrences of each simile marker in a particular novel by the total number of sentences in that novel. That measure however proves itself to not accurately reflect the distribution of verb and adjective simile markers as their use seems to not be influenced by the length of the novel. The frequency of each simile marker concerned in this experiment is therefore measured only by counting its occurrence in a specific novel.

Due to their polysemous use and the noise they introduce in the generated data, markers such as 'remind of' and 'rappeler' were not considered in the final analysis.

The discrepancy between the frequency count of each grammatical category in English and French tends to suggest that both languages work differently as far as simile creation is concerned. In English, verb simile markers appear to be more used than adjective ones. As a matter of fact, two main verb constructions largely surpass the other markers in number: the structures 'seem' + NP and 'be/become a sort/type/kind of…' + NP which count more than 5,000 occurrences.

Excluding adjectives formed by using '-like' or an adjective of colour as a suffix, 'akin (to)' is the most used adjective simile marker with about 350 occurrences and is generally associated with nouns denoting feelings.

As far as French is concerned, the gap between the use of adjective and verb simile markers is less pronounced. Like its English counterpart, 'sembler' has a good place among most frequently used verb markers but is less predominant than 'ressembler (à)'. French writers also distinguish themselves by their preference for introducing similes with the adjectives 'pareil (à)' and 'semblable (à)' .

In addition, in the French corpus, 'similaire (à)' is never used as a typical simile marker, just like 'identical (to)' in the British corpus. If 'identique (to)' appears in the French corpus, it has the smallest number of occurrences of all adjectives that are effectively used to create similes. Similarly, the inflected forms of 'comparable' are found in both corpora but are not so common. A possible explanation as Bouverot (1969) suggests could be the length of the word 'comparable' which surely affects the sentence rhythm.

From the results of this experiment, it is possible to conclude that there seems to exist preferred verb and adjective simile markers in fictional prose. However, even though some writers use them systematically—for example all texts by Balzac present in the corpus contain at least one occurrence of 'pareil à + NP'—the frequency use of these markers in those authors' novels generally vary greatly from one text to another. Consequently, even though verb and adjective simile markers could be interesting stylistic indicators, taken individually, they do not seem to be able to characterise unequivocally one author's style but rather hint at a possible group tendency or an aesthetic idiosyncrasy.

## Acknowledgement

This work was supported by French state funds managed the ANR within the Investissements d'Avenir programme under the reference ANR-11-IDEX-0004-02.

---

[1] www.gutenberg.org
[2] beq.ebooksgratuits.com

*Conference on New Methods in Language Processing*: 44-49.